\documentclass[12pt]{iopart}

\pdfoutput=1
\usepackage[utf8]{inputenc}
\usepackage[USenglish]{babel}

\usepackage{booktabs}
\usepackage{amssymb,amsthm,amsfonts,amsmath}
\usepackage{bbold}
\usepackage{url}
\usepackage[colorlinks=true, allcolors=blue, breaklinks]{hyperref}
\usepackage[linesnumbered,ruled,vlined]{algorithm2e}
\SetKwInput{KwInput}{Input}    
\SetKwInput{KwOutput}{Output}
\usepackage{todonotes}

\usepackage{graphicx}

\usepackage[labelfont=bf, font=footnotesize]{caption}

\makeatletter
\def\l@subsubsection#1#2{}
\makeatother

\def\fig#1{Fig.~\ref{#1}}

\def\0#1#2{\frac{#1}{#2}}  

\graphicspath{{./figures/}}

\begin{document}

\title[Inverting Non-Injective Functions with Twin Neural Network Regression]{Inverting Non-Injective Functions with Twin Neural Network Regression}

\author{Sebastian J. Wetzel}
\ead{sebastian.wetzel@gmx.net}


\begin{abstract}
Non-injective functions are not globally invertible. However, they can often be restricted to locally injective subdomains where the inversion is well-defined. In many settings a preferred solution can be selected even when multiple valid preimages exist or input and output dimensions differ. This manuscript describes a natural reformulation of the inverse learning problem for non-injective functions as a collection of locally invertible problems. More precisely, Twin Neural Network Regression is trained to predict local inverse corrections around known anchor points. By anchoring predictions to points within the same locally invertible region, the method consistently selects a valid branch of the inverse. In contrast to current probabilistic state-of-the art inversion methods, Inverse Twin Neural Network Regression is a deterministic framework for resolving multi-valued inverse mappings. I demonstrate the approach on problems that are defined by mathematical equations or by data, including multi-solution toy problems and robot arm inverse kinematics.

\end{abstract}
\vspace{2pc}
\maketitle

\newpage
\section{Introduction}

Inverse problems are common in science and engineering whenever it is necessary to deduce the input for a given output. Examples include inverse problems in physics~\cite{Kades2020}, chemistry~\cite{Sridharan2022}, engineering~\cite{1993}, computer graphics~\cite{Aristidou2017} or inverse kinematics~\cite{Manjegowda2025}. In many such settings, the forward mapping is known or can be approximated reliably from data, while the inverse mapping is ill-posed or there does not exist a unique solution.

A common obstacle in inverse problems is \emph{non-injectivity} which means that a forward function maps multiple distinct inputs to the same output. In these cases, there is no unique inverse. Traditional regression algorithms are not suitable for providing solutions in these cases. When they are trained to approximate the inverse mapping, they are forced to average over incompatible solutions, often yielding predictions that do not correspond to a valid preimage of the target output.

Inverse problems are often solvable in practice by taking into account additional structure. Smooth non-injective functions are typically locally injective when restricted to suitable sub-domains. Hence, solutions can be obtained restricting the input space. 

\begin{figure*}[t!]
    \centering
    \includegraphics[width=\textwidth]{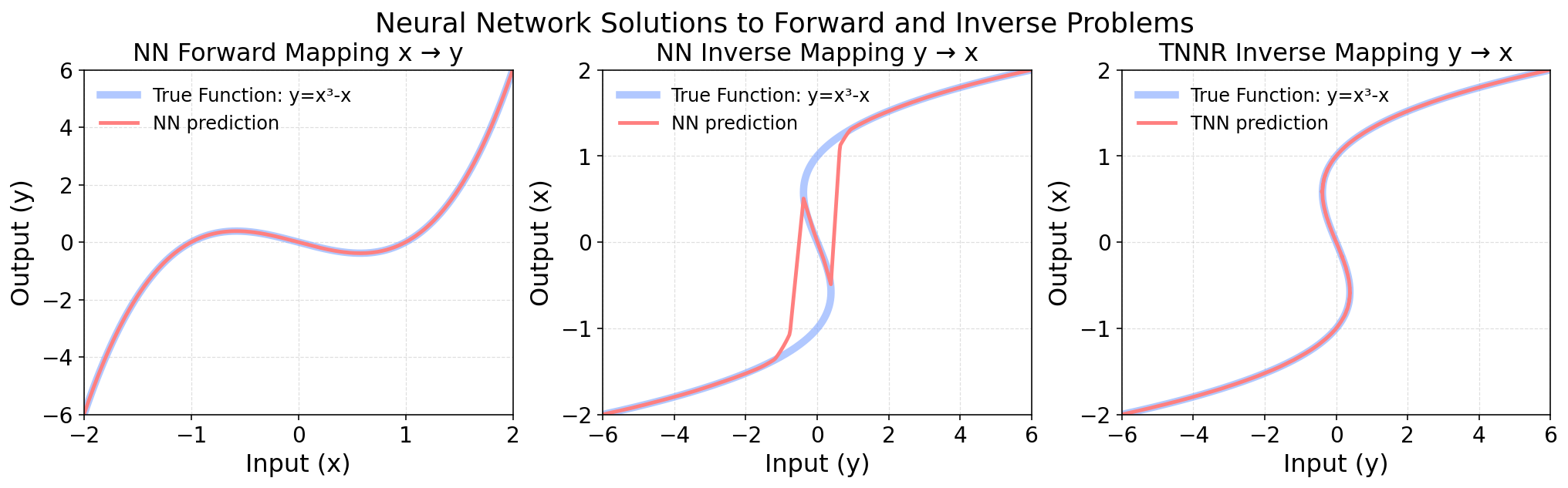}
    \caption{Comparison between forward and inverse approximations of the function $f(x)=x^3-x$ using traditional neural networks and twin neural network regression. Traditional neural networks are capable of approximating the forward function, fail however at retrieving the inverse function because of non-injectivity. In the region where there are multiple preimages for a given target, a traditional neural network becomes confused and attempts to average over predictions. ITNNR can resolve this ambiguity and can hence be employed to fully approximate all branches of the inverse function.}
    \label{fig:combined_prediction_example}
\end{figure*}

In this work, I propose the Inverse Twin Neural Network Regression(ITNNR) method, see figure~\ref{fig:combined_prediction_example}, that can naturally circumvent the problem of non-injectivity. This method is deterministic, does not rely on random sampling of solutions, and can learn approximations to non-linear non-injective continuous functions. It is important to emphasize the deterministic nature, as many probabilistic inversion methods such as mixture models~\cite{bishop1994mixture} suffer from distinct limitations. Although probabilistic inversion can resolve multimodality, many practical and theoretical settings require deterministic inverse mappings that are stable, reproducible, interpretable, computationally efficient, and compatible with other elements in deterministic pipelines. A common example for these requirements is inverse kinematics, especially when there is no known exact forward function which can be optimized with iterative solvers.

The framework is based on Twin Neural Network Regression~\cite{Wetzel2022a} (TNNR) combined with k-nearest neighbor (k-NN) search~\cite{Wetzel2024}. Rather than learning an explicit inverse function, the proposed method predicts local adjustments to known anchor points that share similar output values. This reformulation automatically selects regions of local invertibility and avoids averaging across incompatible branches of the inverse. By anchoring predictions in the vicinity of known solutions, the method can recover multiple valid preimages of a target output.

TNNR is a machine learning method that learns to predict differences between target values instead of the targets themselves. It was originally introduced to improve the regression accuracy in small data problems. TNNR produces an ensemble of predictions from a single trained model by anchoring the prediction to multiple known data points. This property is particularly well suited for inverse problems since different anchors implicitly define different locally injective regions of a non-injective function, enabling the recovery of multiple inverse branches without requiring separate models or explicit branch labels.

By combining TNNR with k-nearest neighbor (kNN) search, the proposed framework selects anchors that define locally consistent inversion regions. In each of these regions, TNNR learns how to translate changes in the target variable into corresponding changes in the input space. 

The proposed method can be applied to approximate pseudoinverses of problems that are (1) defined purely by measured data and to problems where (2) the formula of the forward function is known. In the latter case, the forward function can be used to rank possible candidate solutions, while in the former case the forward functions need to be approximated by a traditional neural network first.

\subsection{Outline}

This paper is organized as follows: Section~\ref{sec:priorwork} reviews prior work on twinned regression methods and related approaches to inverse problems. Section~\ref{sec:tnnr} provides a review of Twin Neural Network Regression. Section~\ref{sec:inversetnnr} introduces the proposed inversion framework and its variants. The underlying mechanisms of the ITNNR framework are demonstrated in section~\ref{sec:toy_problems} at the example of simple toy problems. Section~\ref{sec:experiments} presents experimental results when applying the method to problems defined by data or mathematical functions. Section~\ref{sec:discussion} discusses the results and proposes future directions for application and research.

\section{Prior Work}
\label{sec:priorwork}
The Twin Neural Network Regression (TNNR or input doubling or pairwise difference) framework is a regression technique that is based on learning the differences between target values instead of the regression targets themselves. A regression problem can be solved by aggregating these differences anchored at different known data points.

The neural network formulation of this approach was initially introduced in 2020 as \emph{Twin Neural Network Regression}~\cite{Wetzel2022a} (Submission: 29 Dec 2020, arXiv publication: 29 Dec 2020, Journal publication 04 October 2022). Independently, the same method was introduced as~\emph{Input Doubling Method} for radial basis function (RBF) neural networks~\cite{Izonin2021} (Submission: 29 November 2020, Conference Publication: 21 July 2021). Following these initial developments, the approach was subsequently applied to decision trees as \emph{Pairwise Difference Regression}~\cite{Tynes2021} (Submission: 14 June 2021, Journal Publication: August 4, 2021).

This framework has been extended by several conceptual developments. Belaid et al. proposed a systematic way to identify the best anchors~\cite{BelaidEtAl2024WeightedPDL}. Izonin et al. proposed a method to optimize training time with clustering~\cite{Izonin2024d}. A general examination of the reasons for the outperformance of this framework can be found in~\cite{https://doi.org/10.48550/arxiv.2301.01383}. The mismatch of predictions and the violation of consistency conditions give rise to several uncertainty estimates~\cite{Wetzel2022a,Tynes2021,Belaid2025}. The same consistency conditions can also be employed during training to transform TNNR into a semi-supervised regression algorithm~\cite{Wetzel2022b}. Furthermore, the overall framework can be employed for classification problems, too~\cite{Izonin2024,BelaidEtAl2025PDLClassification}.

Important to note is that TNNR can naturally be combined with k-nearest neighbor (k-NN) search~\cite{Wetzel2024} to increase performance in specific problems. This variant of TNNR forms the basis of the method proposed in this article.

The TNNR framework has been applied in various forms to a wide range of problems. It has shown incredible performance, especially in scenarios with scarce training data, sometimes cutting error measures compared to previous state-of-the-art methods by more than $50\%$.

These applications include medicine~\cite{Izonin2021b,Izonin2021c}, energy efficiency of buildings~\cite{Izonin2024b}, bridge damage estimation~\cite{Izonin2024c}, natural language processing~\cite{xie-etal-2022-automated}, image processing~\cite{Hu2023}, quantum~\cite{Chen2023}, drug discovery\cite{Wang2024,Pillai2023}, agriculture\cite{Li2024}, chemistry and materials science \cite{https://doi.org/10.48550/arxiv.2501.09103,Yatkn2024,Kim2024} authorship analysis~\cite{Corbara2023}or ranking\cite{Hlzer2024}.

While non-injective function inversion is a very common problem, providing comprehensive solutions is still an ongoing research agenda. Specialized solutions are common in the field of inverse kinematics~\cite{Jack1993,https://doi.org/10.48550/arxiv.2205.10837}. There have been several proposals that address this problem in a general context with machine learning. Neural networks can be employed to regularize these problems to select specific solutions and to stabilize the inversion process~\cite{Haltmeier2023,Schwab2019}. NETT replaces hand-crafted regularizers with trained neural networks~\cite{Li2020}. 

Among the most popular methods for function inversion are invertible neural networks and normalizing flows~\cite{https://doi.org/10.48550/arxiv.1808.04730,https://doi.org/10.48550/arxiv.1605.08803}. Further, mixture density networks explicitly model multimodal inverse mappings~\cite{bishop1994mixture,https://doi.org/10.48550/arxiv.2303.15244} by predicting parameters of a Gaussian mixture model conditioned on the input. Since all of these methods are probabilistic, the most reasonable baseline for a general deterministic function inversion is provided by a simple lookup table. This lookup based inversion can be implemented analogously to the k-nearest neighbors (kNN) algorithm~\cite{Cover1967,Fix1989}.

\section{Review of Twin Neural Network Regression}
\label{sec:tnnr}

\begin{figure*}[h!]
    \centering
    \includegraphics[width=0.9\textwidth]{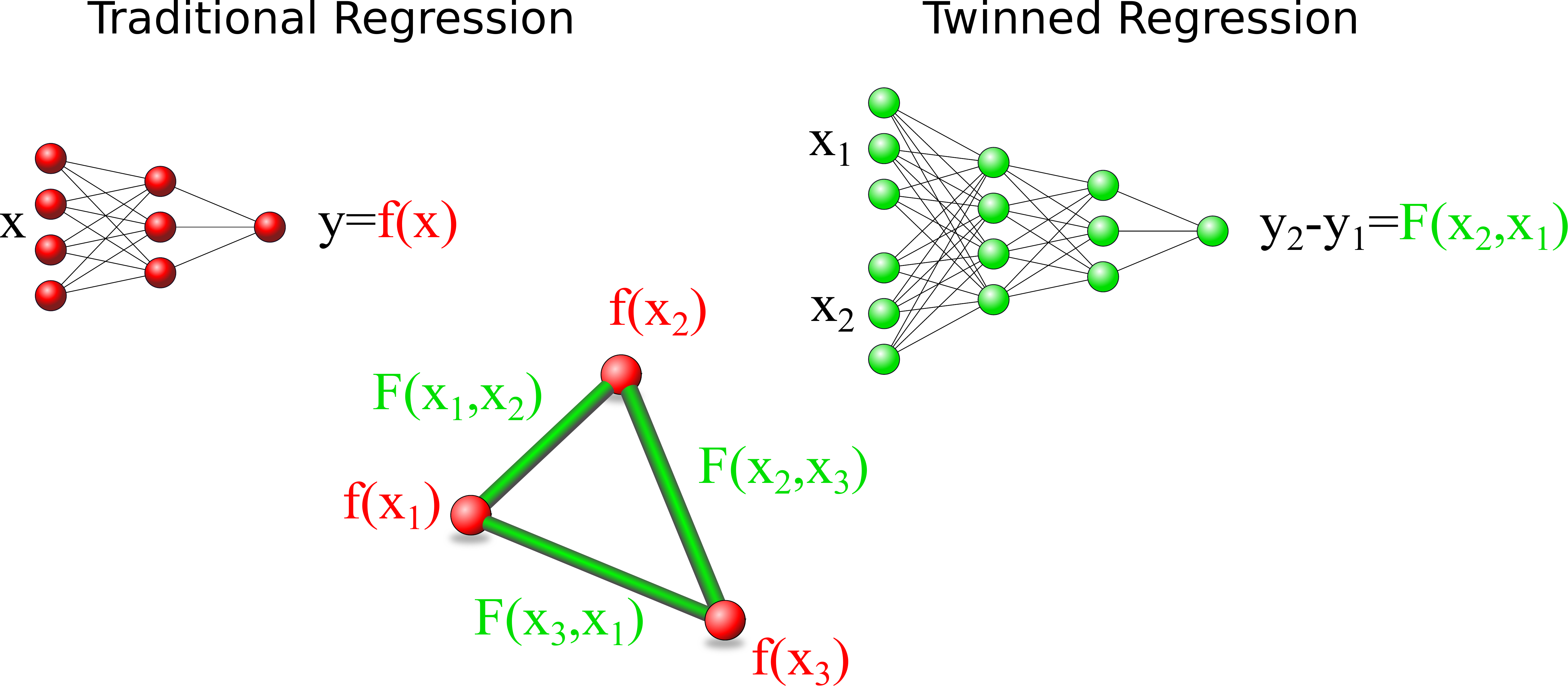}
    \caption{Reformulation of a traditional forward regression problem: A traditional solution to a regression problem consists of finding an approximation to the function that maps a data point $x$ to its target value $f(\mathbf{x})=y$. Twinned regression methods solve the problem of mapping a pair of inputs $x_1$ and $x_2$ to the difference between the target values $F(x_2,x_1)=y_2-y_1$. The resulting function can then be employed as an estimator for the original regression problem $y_2=F(x_2,x_1)+y_1$ given a labelled anchor point $(x_1,y_1)$. Twinned regression methods must satisfy loop consistency: predictions along each loop sum to zero:  $F(x_1,x_2)+F(x_2,x_3)+F(x_3,x_1)=0$.}
    \label{fig:TNNR}
\end{figure*} 

Twin Neural Network Regression (TNNR) is a machine learning method based on predicting differences between the targets of a regression problem instead of the targets themselves. While neural networks are the most powerful base algorithms for a twinned formulation, the base algorithm can also be swapped for many different machine learning algorithms.

Let us consider a regression problem with training data set of $n$ data points $X^{\text{train}}=(\mathbf{x}_1^{\text{train}},...,\mathbf{x}_n^{\text{train}})$ with their corresponding target values $Y^{\text{train}}=(\mathbf{y}_1^{\text{train}},...,\mathbf{y}_n^{\text{train}})$ and a test data set $X^{\text{test}}=(\mathbf{x}_1^{\text{test}},...,\mathbf{x}_m^{\text{test}})$ of size $m$. Both inputs and targets are allowed to exist in higher dimensional spaces $\mathbf{x}_i\in\mathbb{R}^p,\mathbf{y}_i\in\mathbb{R}^q$. To solve this regression problem, one needs to find a function $f$ that minimizes the error between $f(\mathbf{x}_i)$ and $\mathbf{y}_i$ for all unknown data points $\mathbf{x}_i\in X^{\text{test}}$ measured by a predefined objective function. In this work, the final performance is measured by $L_{RMSE}=\sqrt{\sum_{i=1}^n (f(\mathbf{x}_i)- \mathbf{y}_i)^2/n}$ evaluated on test data $(X^{\text{test}},Y^{\text{test}})$.

Traditional regression models aim to model a function that maps input data directly to output targets. However, TNNR reformulates the regression problem into predicting differences between the targets of input data pairs, see \fig{fig:TNNR}. Given two inputs, $\mathbf{x}_i$ and $\mathbf{x}_j$, and their corresponding outputs $\mathbf{y}_i$ and $\mathbf{y}_j$, the network learns to predict
\begin{align}
    F(\mathbf{x}_i,\mathbf{x}_j)= \mathbf{y}_i-\mathbf{y}_j \ .
    \label{eq:TNNR}
\end{align}
By leveraging the relative information between data points, the original regression problem is solved by ensembling the differences between an unknown data point and several known anchor points. The estimated target for a new input $x$ is calculated as:
\begin{align}
    \mathbf{y}^{\text{new}} = \frac{1}{n} \sum_{j=1}^{n} \left( F(\mathbf{x}^{\text{new}}, \mathbf{x}_j^{\text{train}}) +  \mathbf{y}_j^{\text{train}} \right)
    \label{eq:ypred}
\end{align}
This ensembling process is effective in averaging out noise and reducing variance. The finite distance between anchors leads to a high ensemble diversity, causing a significantly improved accuracy in many applications compared to traditional regression frameworks~\cite{Wetzel2022a,Izonin2021,Tynes2021}. Pairing training data points brings the benefit of increasing the effective training data set at the cost of training time, hence TNNR loses effectiveness on data sets of sizes above $10^5-10^6$.

An important feature of TNNR is the requirement for loop consistency. For any closed loop along data points, the sum of the predicted differences should be zero (see \fig{fig:TNNR}):
\begin{align}
&0=F(\mathbf{x}_1, \mathbf{x}_2) + F(\mathbf{x}_2, \mathbf{x}_1)  \\
&0=F(\mathbf{x}_1, \mathbf{x}_2) + F(\mathbf{x}_2, \mathbf{x}_3) + F(\mathbf{x}_3, \mathbf{x}_1) 
    \label{eq:loopc}
\end{align}
The TNNR (or input doubling or pairwise difference) framework is one of the most versatile machine learning techniques because it unifies several properties:
\begin{enumerate}
\item Very high accuracy on many regression problems, especially when training data is scarce.
\item Ability to generate ensembles of predictions by training a single model.
\item The twin training data set can be created by pairing all training data points of the original data set, possibly squaring the effective training data.
\item Consistency conditions allow for estimating prediction uncertainties. 
\item TNNR can be adapted for semi-supervised learning by enforcing consistency on unlabeled data points during the training process ~\cite{Wetzel2022b}.
\end{enumerate}

TNNR (or input doubling or pairwise difference)  offers a powerful and flexible approach to regression problems, particularly in data-scarce environments. Its unique focus on predicting differences and enforcing consistency conditions enables it to achieve high accuracy and robust performance across various applications.

\section{Inverting Non-Injective Functions with Twin Neural Network Regression}
\label{sec:inversetnnr}
\subsection{Overview}

\begin{figure*}[h!]
    \centering
    \includegraphics[width=\textwidth]{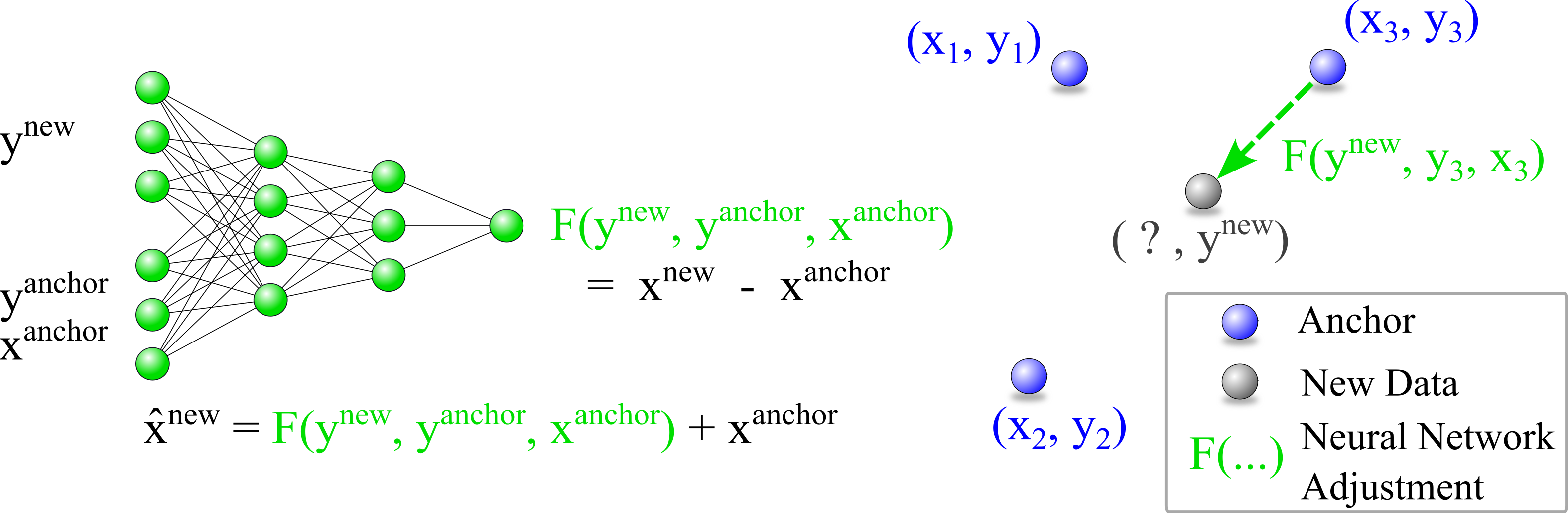}
    \caption{Twin Neural Network Regression for the inverse of non-injective functions: Twin Neural Networks are trained to predict the difference $\mathbf{x}^\text{new}-\mathbf{x}^\text{anchor}$ from inputs $(\mathbf{y}^\text{new},\mathbf{y}^\text{anchor},\mathbf{x}^\text{anchor})$, which is used to estimate the input $\mathbf{x}^\text{new}$ corresponding to the target $\mathbf{y}^\text{new}$. In contrast to TNNR for approximating the forward functions, it is necessary to include a third input variable $\mathbf{x}^\text{anchor}$ because $F$ behaves differently in regions around different anchors $\mathbf{x}_i^\text{anchor}$ that each exhibit the same target $\mathbf{y}^\text{anchor}$.}
    \label{fig:visualization}
\end{figure*}

Non-injective functions are not invertible, since there is no unique input $\mathbf{x}$ for each output $\mathbf{y}$. However, non-injective functions can be restricted to sub-domains on which they are locally injective and surjective and thus invertible if the dimensionality between input and output spaces are the same. Further, even if the dimensionalities do not match it is often possible to choose a preferred solution from many possible solutions. 

In figure \ref{fig:combined_prediction_example} one can see the capabilities of traditional neural networks when it comes to approximating the function $f=x^3-x$. The neural network is successful in replicating the forward function, however, it fails at the attempt of solving the inverse problem. The traditional neural network gets confused by different conflicting $\mathbf{x}$ outputs for each input $\mathbf{y}$. 

I propose a strategy based on twin neural network regression (TNNR) that is capable of inverting this function successfully, denoted inverse TNNR (ITNNR). It is based on predicting adjustments to known anchors $(\mathbf{x}^\text{anchor},\mathbf{y}^\text{anchor})$ such that the predicted $\mathbf{x}^\text{new}$ is one of the possibly many preimages of a given $\mathbf{y}^\text{new}$.
In contrast to TNNR for approximating the forward functions, it is necessary to include a third input variable $\mathbf{x}^\text{anchor}$ because $F$ behaves differently in regions around different anchors $\mathbf{x}_i^\text{anchor}$ that each exhibit the same target $\mathbf{y}^\text{anchor}$, see figure \ref{fig:visualization}. 
The neural network $F$ is trained on a data set of pairs $((\mathbf{x}_i^{\text{train}},\mathbf{y}_i^{\text{train}}),(\mathbf{x}_j^{\text{train}},\mathbf{y}_j^{\text{train}}))$ where the members of each pair are close to each other with respect to their $\mathbf{x}$ coordinates. This can be facilitated by determining the k nearest neighbors of each training data point within the training set. If one has access to the mathematical function that should be inverted, it can also be achieved by perturbing each datapoint within a small but finite range that corresponds to the average distance between anchors. $F$ is trained such that it approximates the differences between the $\mathbf{x}$ coordinates
\begin{align}
    F(\mathbf{y}_i^{\text{train}}, \mathbf{y}_j^{\text{train}},\mathbf{x}_j^{\text{train}})=\mathbf{x}_i^{\text{train}}-\mathbf{x}_j^{\text{train}} \ .
    \label{eq:Ftrain}
\end{align}
A fully trained network $F$ can then be employed to calculate a possible preimage $\mathbf{x}^{\text{new}}$ of $\mathbf{y}^{\text{new}}$ by applying it to an anchor via the formula

\begin{align}
    \hat{\mathbf{x}}^{\text{new}} =F(\mathbf{y}^{\text{new}}, \mathbf{y}^{\text{anchor}},\mathbf{x}^{\text{anchor}})+\mathbf{x}^{\text{anchor}} \ .
    \label{eq:xpred}
\end{align}
The anchors are the nearest neighbors $\mathbf{y}^{\text{new}}$ from a set of known anchors, which is typically the training set. It is important to emphasize that during the training phase neighbors are chosen based on $\mathbf{x}$, but during inference neighbors are chosen based on $\mathbf{y}$. If you have access to the mathematical expression of the forward function you can in principle create infinitely many anchors. However, the inference time will then be limited by the scaling of nearest neighbor search: for a $m$ unknown points, their $k$ nearest neighbors from within an anchor set of size $n$, can be found in $O(n\log n+m(\log n+k))$ steps \cite{10.5555/645924.671192}. That compares unfavorably with neural network inference of $ O(1) $. Conversely, if we had infinitely many anchors which we could search through efficiently, we would not need ITNNR.

\subsection{Inversion Based on Mathematical Formula}
\label{sec:mathematical_formula}
ITNNR takes on a different form based on the setting that it is employed. The two versions are based on the nature of the data. If one has access to the exact mathematical formula one intends to invert, it is possible to create an infinite amount of training data. Furthermore, this data is noiseless and anchors can be arbitrarily created. Creating a pairwise training set involves pairing each training data point with the $k$ closest neighbors in $\mathbf{x}$-space from the original training data set
\begin{align}
D_{Y}&=(\mathbf{y}_i^{\text{train}},\mathbf{y}_j^{\text{train}},\mathbf{x}_j^{\text{train}})_{i,j}\\
D_{X}&=(\mathbf{x}_i^{\text{train}}-\mathbf{x}_j^{\text{train}})_{i,j}
\end{align}
According to the equation \ref{eq:Ftrain} we train a neural network $F$. 
During inference for each new data point $(\ \cdot \ , \mathbf{y}^{\text{new}})$ we pick the $k$ nearest neighbors in $\mathbf{y}$-space from the training set as anchors. For every anchor we predict a preimage $\mathbf{x}^{\text{new}}$ according to equation \ref{eq:xpred}.

The availability of the forward function allows for the projection of the error $\left \|f(\hat{\mathbf{x}}^{\text{new}} )-\mathbf{y}^\text{new}\right \|$ into $\mathbf{y}$-space, avoiding the ambiguity that stems from non-injectivity. This error is available during the inference phase for every new data point. Hence, it can be accessed to rank the predictions
, which is necessary to stay on the correct branches during branch transitions depicted in figure \ref{fig:branch_selection}.

\subsection{Inversion Based on Measured Data}
\label{sec:measured_data}
ITNNR takes on a slightly different form when applied to measurement data. This data is naturally finite and typically noisy. The paired training set obtained by pairing each training data point with the $k$ closest neighbors is $k\times$ larger than the original data set. This fact makes the method very efficient for small data sets. Further, it is necessary to train a traditional neural network to solve the forward problem. This network can then be used instead of a mathematical formula to ensure staying on the correct branch during branch transitions.

Beyond the scope of this manuscript, it might be possible to choose optimal anchors to avoid anchors that contain too much noise, as proposed in~\cite{BelaidEtAl2024WeightedPDL}. 

\subsection{Inverse Twin Neural Network Regression Algorithm}

The complete procedure can be summarized concisely in a two-step algorithm. The training phase consists of the pairwise training of the twin neural network $F$ to predict differences between neighbors, see algorithm~\ref{alg:1}. During the inference phase, a nearest neighbor search algorithm selects anchors that exhibit a nearby target variable $\mathbf{y}$. All of these anchors suggest predictions for valid preimages $\mathbf{x}$ which get ranked by checking for consistency with the forward function, see algorithm~\ref{alg:2}.

\begin{algorithm}
\caption{Inverse Twin Neural Network Regression Training}\label{alg:1}
  \KwData{Labelled data set $(X^\text{train},Y^\text{train})$} 
  \KwInput{Number of nearest anchors  $k$}

  \SetKwFunction{FMain}{Neighbors}
  \SetKwProg{Fn}{Function}{:}{}
  \Fn{\FMain{$\mathbf{x}$, $X^\text{ref}$, $k$}}{
            $L$ $\leftarrow$ select k nearest neighbors of $\mathbf{x}$ within $X^\text{ref}$\\
        \KwRet $L$
  }

  $D_Y\leftarrow[(\mathbf{y}_i,\mathbf{y}_j,\mathbf{x}_j) $
  for $ \mathbf{x}_j \in \FMain(\mathbf{x}_i,X^\text{train},k)$
  for $ \mathbf{x}_i \in X^\text{train}  ]$\\
$D_X\leftarrow[(\mathbf{x}_i-\mathbf{x}_j) $
  for $ \mathbf{x}_j \in \FMain(\mathbf{x}_i,X^\text{train},k)$
  for $ \mathbf{x}_i \in X^\text{train}  ]$\\

  initialize twin neural network regression model $F$\\
  train $F$ on $(D_Y,D_X)$\\

  \KwOutput{Trained model $F$}  
\end{algorithm}

\begin{algorithm}
\caption{Inverse Twin Neural Network Regression Inference}\label{alg:2}
  \KwData{ Labelled data set $(X^\text{anchor},Y^\text{anchor})$\\ \phantom{\textbf{Data: }} New data target $y^\text{new}$} 
  \KwInput{Trained twin neural network regression model $F$\\
  \phantom{\textbf{Input: }}Forward function $f$\\
  \phantom{\textbf{Input: }}Number of nearest anchors  $k$}

  \SetKwFunction{FMain}{Neighbors}
  \SetKwProg{Fn}{Function}{:}{}
  \Fn{\FMain{$y$, $Y^\text{ref}$, $k$}}{
            $L$ $\leftarrow$ select k nearest neighbors of $y$ within $Y^\text{ref}$\\
        \KwRet $L$
  }

  $D_Y^\text{inference}\leftarrow[(\mathbf{y}^\text{new},\mathbf{y}_j,\mathbf{x}_j) $
  for $ y_j \in \FMain(y^\text{new},Y^\text{anchor},k)]$\\
$X^\text{inference}\leftarrow[\mathbf{x}_j $
  for $ y_j \in \FMain(y^\text{new},Y^\text{anchor},k)]$\\
  
  List of predictions by anchors: $P\leftarrow ( F(D_Y^\text{inference})+X^\text{inference})$

Sort predictions by consistency: $P^\text{sorted}\leftarrow$ sort $P$ by increasing $\left \|f(P)-\mathbf{y}^\text{new}\right \|$\\

  \KwOutput{Predictions $P^\text{sorted}$}   		
\end{algorithm}

\begin{figure*}[h!]
    \centering
    \includegraphics[width=0.6\textwidth]{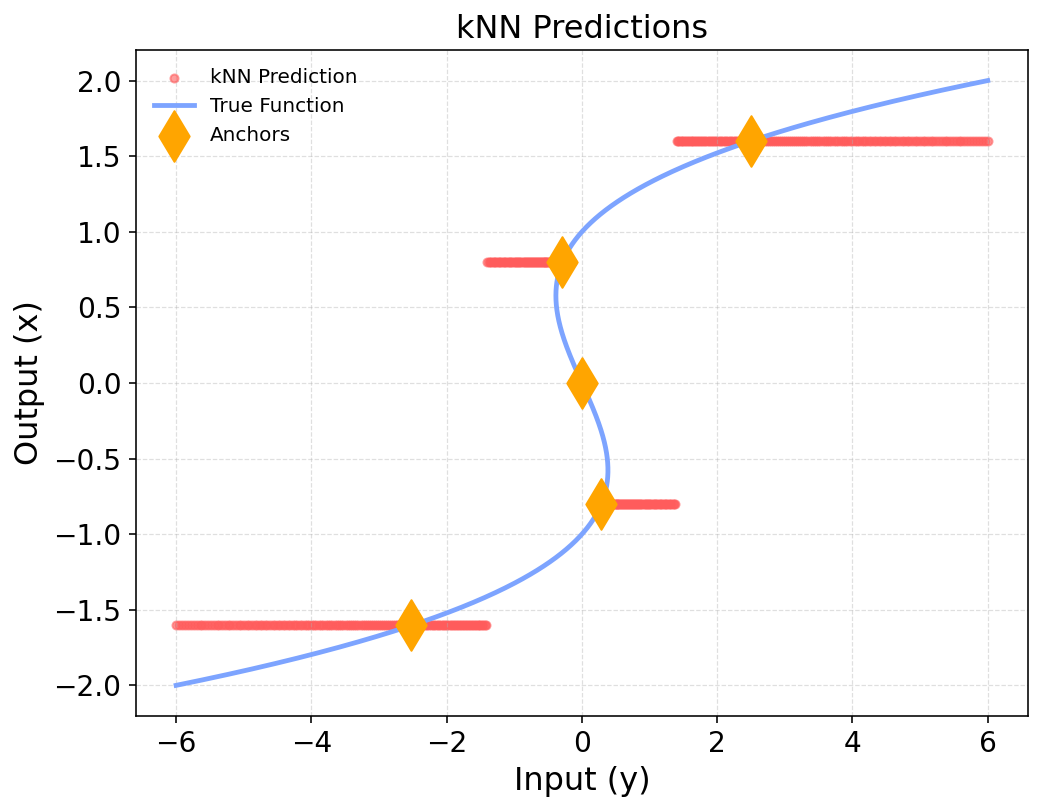}
    \caption{Employing the k-nearest-neighbors(kNN) algorithm to find the inverse of the function $f(x)=x^3-x$. The distance that kNN uses to choose the closest neighbor is the absolute value of the difference in $y$. Since non-injectivity prevents averaging of kNN predictions, it is necessary to choose to make predictions based solely on one anchor each time. The 5 anchors define 5 different preimages. The image contains a very unfavorable choice of anchors, highlighting the problems of choosing kNN to find the inverse of functions. The lack of sufficient anchors leads to the assignment of wrong branches to make predictions. Increasing the number of anchors substantially will lead to an image similar to the right in figure \ref{fig:combined_prediction_example}.}
    \label{fig:knn_predictions}
\end{figure*}
\section{Toy Problems}
\label{sec:toy_problems}

\subsection{Baseline k-Nearest Neighbors Problems}
Inverting a function via a lookup table provides a stable baseline for examining the ITNNR method. It can be implemented with the k-nearest neighbor algorithm. Applying this algorithm to the inversion of the function $f(x)=x^3-x$ is depicted in figure \ref{fig:knn_predictions}. 

By artificially setting the number of anchors to 5 it is possible to visualize the problems of assigning the correct branches during function inversion. It is important to note that this number of branches is of course far lower than any suitable training set, and far below any practical limit by computing power vs scaling. 

One can see that not a single branch is captured accurately. Furthermore, in the region of non-injectivity, the nearest neighbor search picks up the anchors from the wrong branch, leading to a discontinuous inverse prediction. Both of these problems can be avoided by increasing the number of training data points to infinity or by employing ITNNR.

It is important to note that the nearest neighbor search algorithm cannot be allowed to average over predictions, since the averaging process that is inherent to kNN with $k>1$ mixes different preimages. However, it is possible to obtain independent predictions from other neighbors. Trivially, kNN with $k=1$ is equivalent to ITNNR using only nearest neighbor anchors, if $F=0$. For these reasons, kNN is a good baseline for all experiments.

\begin{figure*}[ht!]
    \centering
    \includegraphics[width=0.6\textwidth]{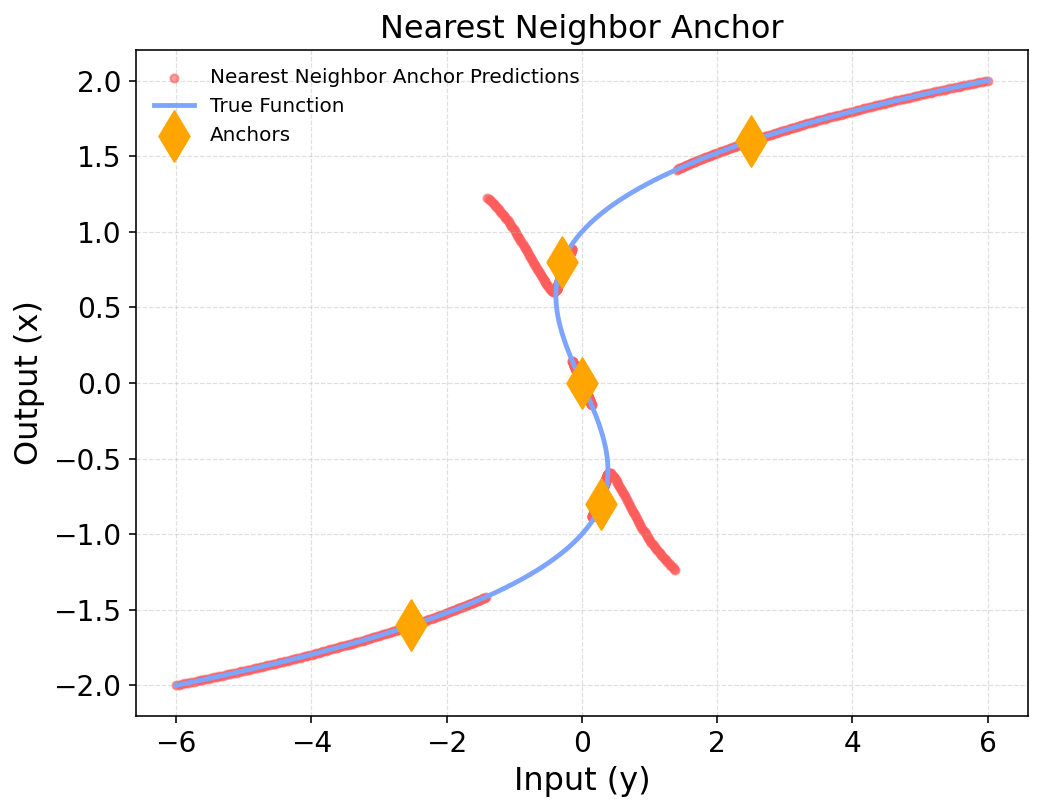}
    \caption{A first attempt at employing TNNR to solve find the inverse of the function $f(x)=x^3-x$. The anchors are chosen as nearest neighbors based on distance in $y$. While TNNR with nearest neighbor anchors has similar flaws as kNN, one can clearly see that some branches are very well approximated, even in the region where the function is non-injective. The lack of sufficient anchors still leads to the assignment of wrong branches to make predictions. Increasing the number of anchors from 5 to 600 corresponds to the TNNR prediction in image \ref{fig:combined_prediction_example}.}
    \label{fig:nearestneighboranchors}
\end{figure*}
\subsection{Inverse Twin Neural Network Regression.}

Let us now employ ITNNR to approximate the inverse of the function $f(x)=x^3-x$.
We consider the same scenario as in the previous kNN section. Most importantly, there are only 5 very unfavorably placed anchors. To demonstrate the power of ITNNR we train $F$ with a generator that samples training pairs. The prediction of the preimages is anchored on the nearest anchors based on the distance in $\mathbf{y}$. Figure \ref{fig:nearestneighboranchors} displays the results. We can see that in the vicinity of all anchors the shape of the function is approximated correctly. Moreover, one can clearly see that ITNNR learns a non-linear approximation of the function over a finite distance, instead of just a linear residual correction around each anchor. However, the result still suffers from the wrong assignment of branches in the vicinity of extrema of the forward function. Further, since we only used one anchor for each $\mathbf{y}$ there are obviously some gaps in the approximation.

Both, the gaps in the prediction and the problems at the extrema of the forward functions can be resolved by increasing the number of anchors as seen in the TNNR prediction in figure \ref{fig:combined_prediction_example}.

\begin{figure*}[ht!]
    \centering
    \includegraphics[width=\textwidth]{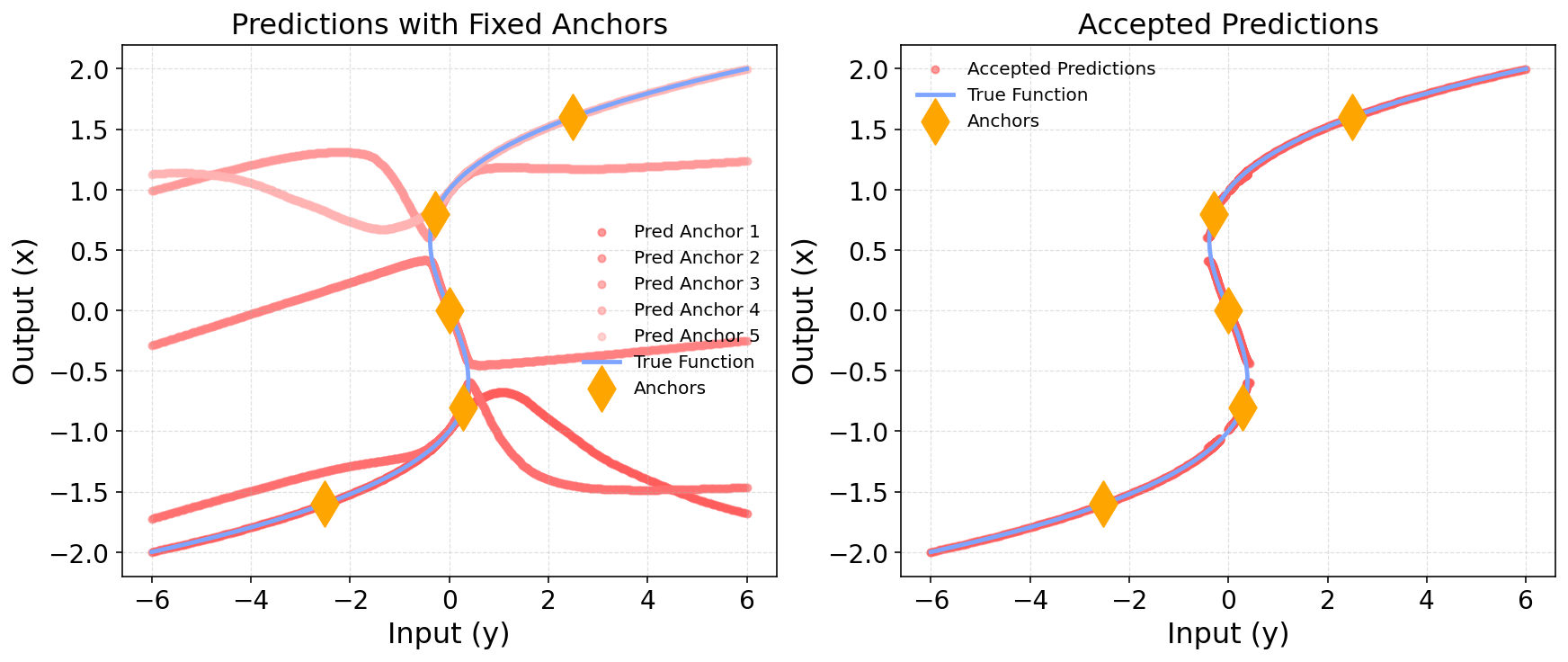}
    \caption{A successful inversion of the function $f(x)=x^3-x$ using TNNR with a very small number of anchors. On the left we can see that each anchor predicts branches that exhibit a piecewise overlap with the true function. These branches can be separated by enforcing consistency via the forward function to create the final prediction as seen on the right. If this function is unknown it can be obtained by approximating the forward problem with a traditional neural network. An observant reader might spot gaps in the prediction of the inverse. These stem from the small number of anchors applied to 1000 different $y$ values in regions where the slope of the inverse function is very steep, hence the predictions are farther apart.  }
    \label{fig:branch_selection}
\end{figure*}

However, ITNNR does not rely on this increase of the number of anchors. There is another solution which relies on combining the predictions from multiple anchors. Let us look at the predictions from all 5 anchors over the whole domain in figure \ref{fig:branch_selection}. We can see that most predictions only agree with the function in a small subregion. However, combining all predictions, there is always, at every $\mathbf{y}$, at least one predicted branch that overlaps with the true function.

If we have access to the exact mathematical forward function $f$ as described in section \ref{sec:mathematical_formula}, we can use it to select which predicted branches correspond to the true function. For each predicted $\mathbf{x}$ it is possible to compare $f(\mathbf{x})$ with the input $\mathbf{y}$. If they match, we accept the branch, if not, we reject it. If the forward function is not available as it is the case with measurement data, as outlined in section \ref{sec:measured_data}, we can still train a traditional neural network to take on the role of the forward function. Even if this function is inaccurate, the function can still serve as a branch selector, since wrong branches strongly deviate from the function. 

Let us now employ the branch selection strategy in figure \ref{fig:branch_selection}. We can see that even with a very small number of anchors, the preimages of the function can be faithfully recovered. The predictions cover the whole domain with the exception of some minor gaps.

\subsection{Resolving Infinite Solutions Problem}
\label{sec:infinite_solutions}
\begin{figure*}[h!]
    \centering
    \includegraphics[width=0.7\textwidth]{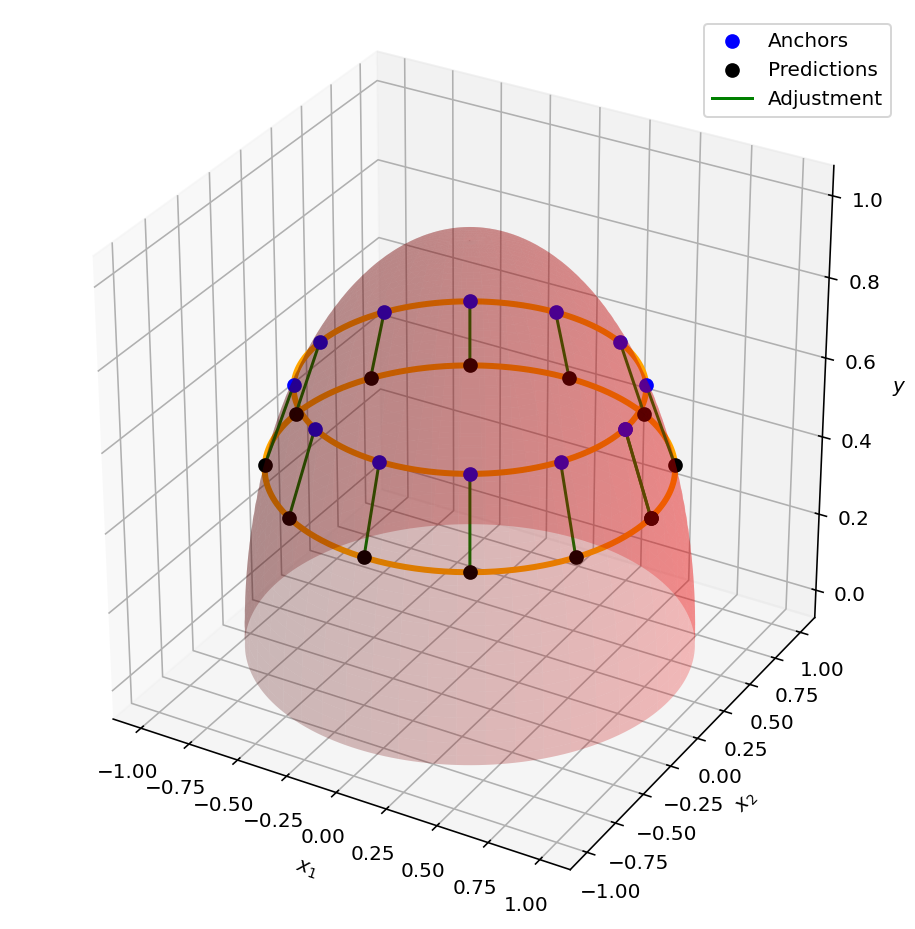}
    \caption{Twin Neural Network Regression for the inverse of non-injective functions where input and output dimensions do not match: A half ball can be parametrized as two branches $y=f(x_1,x_2)=\sqrt{1-x_1^2-x_2^2}$. For each anchor in the vicinity of a desired target $y$ TNNR can find the corresponding preimages $(x_1,x_2)$. Hence, by choosing suitable anchors it is possible to bias the location of the preimage towards preferred regions of the input space.}
    \label{fig:ball}
\end{figure*}

A particularly challenging class of inverse problems arises when there are not just a finite number of preimages for a target value but infinitely many preimages. This situation typically occurs when the dimensionality of the input space exceeds that of the output space. A simple illustrative example is shown in figure~\ref{fig:ball}. Consider the mapping
\begin{align}
    y = f(x_1,x_2) = \sqrt{1 - x_1^2 - x_2^2},
\end{align}
which parameterizes the upper half of the unit ball. For any fixed target value $y \in (0,1)$, the set of preimages $(x_1,x_2)$ forms a circle. Hence, there are infinitely many valid solutions, all of which are equally consistent with the forward function. 

ITNNR resolves this ambiguity by explicitly anchoring the inverse prediction to a known anchor point $(\mathbf{x}^{\text{anchor}},\mathbf{y}^{\text{anchor}})$. More precisely, for a given target $\mathbf{y}^{\text{new}}$, the network predicts an adjustment which yields a candidate preimage
\begin{align}
    \hat{\mathbf{x}}^{\text{new}} = F(\mathbf{y}^{\text{new}}, \mathbf{y}^{\text{anchor}}, \mathbf{x}^{\text{anchor}})+\mathbf{x}^{\text{anchor}}
\end{align}
In the half-ball example, anchors located at different positions on each circular preimage manifold will lead to different inverse solutions, all of which satisfy the forward constraint $f(\hat{\mathbf{x}}^{\text{new}})=\mathbf{y}^{\text{new}}$. From this perspective, ITNNR does not attempt to retrieve the full infinite set of preimages. Instead, given a target output and a set of anchors, ITNNR finds one locally consistent preimage associated with each anchor. 

Importantly, the choice of anchors provides a natural mechanism for encoding preferences or constraints. By selecting anchors from specific regions of the input space, one can bias the inverse solution toward the desired configurations. For example, in the case of inverse kinematics, the anchor can be chosen as the current state, and ITNNR predicts the minimal joint motion towards a new state. In this sense, anchors play a role analogous to gauge fixing or coordinate selection: they do not eliminate the underlying ambiguity of the inverse problem, but they make a particular solution branch explicit and reproducible.

\section{Experiments}
\label{sec:experiments}

This section examines the application of the proposed ITNNR framework on a diverse set of inverse problems. The experiments are designed to assess the ability of ITNNR to recover valid preimages of non-injective functions under varying dimensionalities, levels of nonlinearity, and noise conditions. Performance is measured using the root mean squared error (RMSE) in target space between $f(\hat{\mathbf{x}})$ and the true input $\mathbf{y}$, where $f$ is the exact forward function (even in the case of noisy data to guarantee the most accurate performance measurement).
This performance is compared against deterministic machine learning methods, k-nearest neighbor (kNN) inversion and traditional neural networks (NN) and probabilistic mixture density networks (MDN). The experiments mimic two scenarios: at first, we assume the exact forward function is known, this allows for the generation of infinitely many noisy free data points and for direct consistency checking. Second, real data is typically limited and noisy. In these cases, the forward function for consistency checking needs to be approximated by a traditional neural network.

\subsection{Data}

The inversion problems analyzed in this section are summarized below and described in detail in \ref{sec:data}:

\begin{itemize}
\item Exp 1: 1D $\rightarrow$ 1D cubic polynomial 
\item Exp 2: 1D $\rightarrow$ 1D quartic polynomial
\item Exp 3: 2D $\rightarrow$ 1D upper half of the unit ball
\item Exp 4: 2D $\rightarrow$ 2D bivariate polynomial
\item Exp 5: 3D $\rightarrow$ 3D trivariate polynomial
\item Exp 6: 2D $\rightarrow$ 2D planar 2-link robot arm
\item Exp 7: 3D $\rightarrow$ 2D planar 3-link robot arm
\item Exp 8: 3D $\rightarrow$ 3D yaw–pitch–pitch robot arm
\item Exp 9: 6D $\rightarrow$ 3D six degree-of-freedom robot arm
\end{itemize}

For experiments 1-3, 500 data points are uniformly sampled from the input domain, with 300 points used for training and as anchors during inference. Of the remaining 200 data points, 100 are used as a validation set and 100 as a test set. For experiments 4–9, all these numbers are doubled. Note that in general, the anchor set can be different from the training set. In experiments where the exact forward function is known, every training batch includes newly sampled data points through a data generator, hence allowing for an unlimited amount of training data.

All experiments involving known forward functions use the exact analytical expression during inference to rank inverse candidates. In the noisy setting, an additional neural network is trained to approximate the forward mapping and is used as a surrogate consistency check, as described in Section~\ref{sec:mathematical_formula}.

\subsection{Experimental Setup}

All neural networks (traditional regression, MDN and TNNR) share the same base architecture and training procedure, which is described in \ref{sec:nn_architecture}. The hyperparameters are kept fixed across experiments, with only minor adjustments to accommodate differences in input and output dimensionality. In several experiments MDN was unstable requiring manual hyperparameter optimization. During training and inference we limit all algorithms to the 5 nearest neighboring anchors, while the generator samples pairs according to the distance between the 5 nearest neighbors. Each experiment is repeated 5 times with a different random seed to estimate the mean and standard deviation of the final RMSE.

\subsection{Results with Known Forward Function}

\begin{table}[t]
\centering
\caption{Results of inversions of non-injective functions with known forward function. The performance is measured in RMSE (mean $\pm$ std) projected in $\mathbf{y}$ space. Traditional neural networks (NNs) fail to converge to a meaningful solution. Nearest neighbors inversion (kNN) performance is far behind inverse twin neural network regression (ITNNR). Mixture density networks (MDN) significantly outperform kNN but can not compete with ITNNR. The by far best result is achieved with ITNNR after branch selection by enforcing consistency through projecting into $\mathbf{y}$ space via the forward function. }\label{tab:clean_results}
\begin{tabular}{lccccc}
\toprule
 & kNN & NN & MDN & ITNNR & Best ITNNR \\
 & 1st anchor &  &  & 1st anchor & (exact forward) \\
\midrule
Exp 1 & $0.074 \pm 0.000$ & $0.122 \pm 0.017$ & $0.011 \pm 0.006$ & $0.008 \pm 0.003$ & $\mathbf{0.005 \pm 0.001}$ \\
Exp 2 & $0.018 \pm 0.000$ & $0.311 \pm 0.004$ & $0.010 \pm 0.001$ & $0.003 \pm 0.001$ & $\mathbf{0.003 \pm 0.000}$ \\
Exp 3 & $0.004 \pm 0.000$ & $0.398 \pm 0.006$ & $0.037 \pm 0.011$ & $0.002 \pm 0.001$ & $\mathbf{0.001 \pm 0.000}$ \\
Exp 4 & $1.063 \pm 0.000$ & $3.961 \pm 0.078$ & $0.254 \pm 0.028$ & $0.399 \pm 0.010$ & $\mathbf{0.153 \pm 0.006}$ \\
Exp 5 & $1.104 \pm 0.000$ & $2.141 \pm 0.060$ & $0.164 \pm 0.017$ & $0.224 \pm 0.004$ & $\mathbf{0.089 \pm 0.003}$ \\
Exp 6 & $0.053 \pm 0.000$ & $0.230 \pm 0.004$ & $0.011 \pm 0.001$ & $0.009 \pm 0.001$ & $\mathbf{0.004 \pm 0.000}$ \\
Exp 7 & $0.120 \pm 0.000$ & $0.624 \pm 0.007$ & $0.073 \pm 0.009$ & $0.025 \pm 0.002$ & $\mathbf{0.013 \pm 0.001}$ \\
Exp 8 & $0.135 \pm 0.000$ & $0.214 \pm 0.003$ & $0.042 \pm 0.010$ & $0.030 \pm 0.001$ & $\mathbf{0.012 \pm 0.001}$ \\
Exp 9 & $0.058 \pm 0.000$ & $0.087 \pm 0.001$ & $0.047 \pm 0.002$ & $0.016 \pm 0.001$ & $\mathbf{0.007 \pm 0.000}$ \\
\bottomrule
\end{tabular}
\end{table}
\begin{table}[t]
\centering
\caption{Results of inversions of non-injective functions defined through noisy data. The performance is measured in RMSE (mean $\pm$ std) projected in $\mathbf{y}$ space. Traditional NNs fail to converge to a meaningful solution. kNN performance is far behind ITNNR. MDNs are on par with ITNNR based on nearest neighbor anchors. However, they fall behind ITNNR after branch selection by enforcing consistency through projecting into $\mathbf{y}$ space via the learned forward function. For comparison, ITNNR results with exact consistency checks are included (in practice this is unavailable since the forward function is not explicitly known). }\label{tab:noisy_results}
\begin{tabular}{lccccc}
\toprule
 & kNN & MDN & ITNNR & Best ITNNR  & Best ITNNR  \\
 & 1st anchor &  & 1st anchor & (learned forward) & (exact forward) \\
\midrule
Exp 1 & $0.086 \pm 0.000$ & $0.020 \pm 0.008$ & $0.024 \pm 0.000$ & $\mathbf{0.017 \pm 0.001}$ & $0.006 \pm 0.000$ \\
Exp 2 & $0.032 \pm 0.000$ & $0.017 \pm 0.003$ & $0.021 \pm 0.000$ & $\mathbf{0.013 \pm 0.001}$ & $0.008 \pm 0.000$ \\
Exp 3 & $0.015 \pm 0.002$ & $0.067 \pm 0.016$ & $0.014 \pm 0.002$ & $\mathbf{0.009 \pm 0.001}$ & $0.004 \pm 0.000$ \\
Exp 4 & $1.056 \pm 0.000$ & $0.406 \pm 0.087$ & $0.453 \pm 0.030$ & $\mathbf{0.243 \pm 0.009}$ & $0.217 \pm 0.014$ \\
Exp 5 & $1.109 \pm 0.000$ & $0.163 \pm 0.010$ & $0.298 \pm 0.006$ & $\mathbf{0.151 \pm 0.002}$ & $0.138 \pm 0.003$ \\
Exp 6 & $0.054 \pm 0.000$ & $0.013 \pm 0.001$ & $0.019 \pm 0.000$ & $\mathbf{0.011 \pm 0.000}$ & $0.009 \pm 0.000$ \\
Exp 7 & $0.121 \pm 0.000$ & $0.058 \pm 0.008$ & $0.040 \pm 0.001$ & $\mathbf{0.025 \pm 0.001}$ & $0.019 \pm 0.000$ \\
Exp 8 & $0.136 \pm 0.000$ & $0.041 \pm 0.006$ & $0.043 \pm 0.001$ & $\mathbf{0.025 \pm 0.001}$ & $0.021 \pm 0.001$ \\
Exp 9 & $0.059 \pm 0.000$ & $0.045 \pm 0.002$ & $0.047 \pm 0.000$ & $\mathbf{0.028 \pm 0.000}$ & $0.026 \pm 0.000$ \\
\bottomrule
\end{tabular}
\end{table}

\begin{figure*}[h!]
    \centering
    \includegraphics[width=0.85\textwidth]{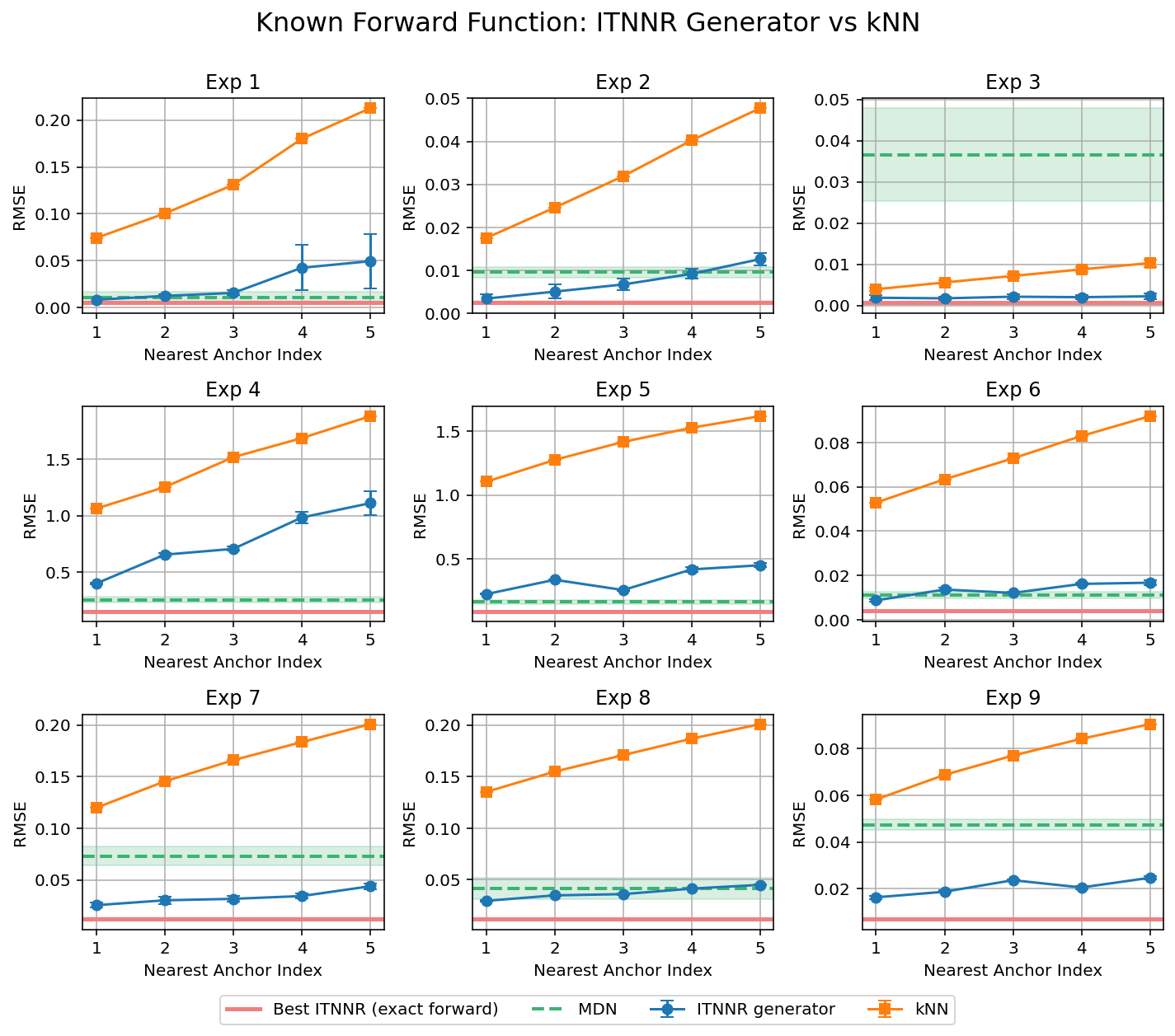}
    \caption{Results of inversions of non-injective functions with known forward function. The performance is measured in RMSE projected in $\mathbf{y}$ space. Performance decreases the farther the anchor is away, however ITNNR results accuracy decays slower than kNN. MDNs lack behind ITNNR. The by far best result is achieved with ITNNR with branch selection by enforcing consistency through projecting into $\mathbf{y}$ space via the known forward function. }
    \label{fig:results}
\end{figure*}
\begin{figure*}[h!]
    \centering
    \includegraphics[width=0.85\textwidth]{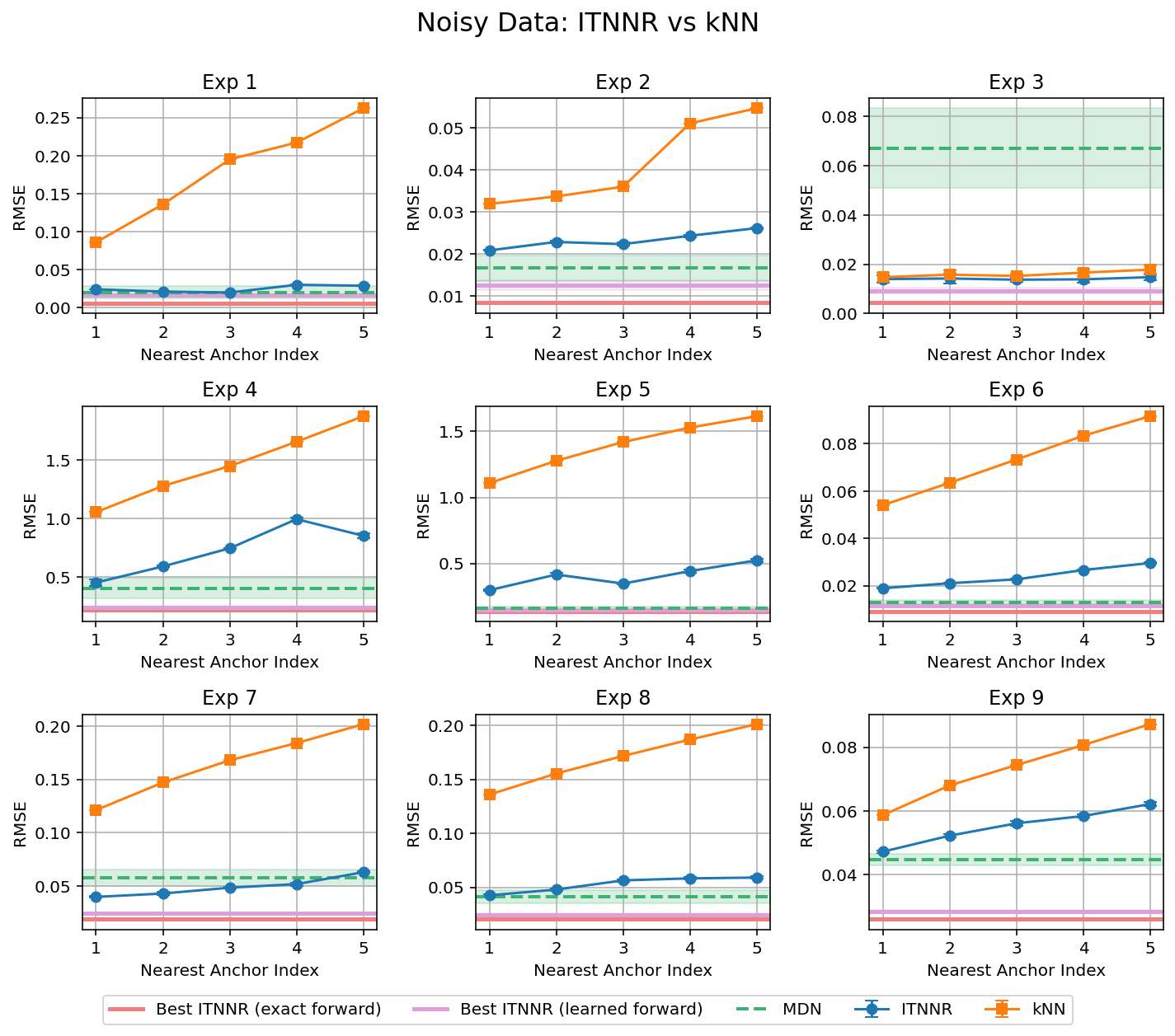}
    \caption{Results of inversions of non-injective functions from noisy data. The performance is measured in RMSE projected in $\mathbf{y}$ space. Performance decreases the farther the anchor is away, however ITNNR results accuracy decays slower than kNN. MDNs can compete with nearest neighbor anchor ITNNR. The by far best result is achieved with ITNNR and branch selection via the learned forward function. For comparison, consistency checks with the exact forward function are included, too. }
    \label{fig:results_noisy}
\end{figure*}

Figure~\ref{fig:results} visualizes representative inversion results across all experiments for the case when the forward function is explicitly known. Quantitative performance is summarized in Table~\ref{tab:clean_results}.

Across all nine experiments, ITNNR significantly outperforms all baselines. Traditional neural networks trained directly on the inverse mapping fail consistently due to non-injectivity and often produce predictions that lie between incompatible solution branches. kNN inversion performs better than neural network inversion but suffers from branch switching and poor local accuracy. MDNs consistently outperform kNN in all experiments except in experiment 3. MDNs perform well on problems with a constant small number of preimages since they assume a finite mixture of modes. For problems such as the half-ball, the inverse manifold is continuous and cannot be represented with a finite mixture without severe approximation error.

ITNNR using a single anchor already yields large error reductions relative to kNN, demonstrating the benefit of learning nonlinear local inverse corrections compared to a lookup table. After branch selection, ITNNR achieves the best performance in all experiments, with RMSE reductions between $80\%$ and $93\%$ compared to kNN and between $46\%$ and $83\%$ compared to MDN (excluding experiment 3).

These improvements are consistent across one-dimensional toy problems and higher-dimensional inverse kinematics tasks, indicating that the method scales robustly to more complex problems.

\subsection{Results with Noisy Data}

To evaluate robustness under realistic conditions, multiplicative Gaussian noise of $1\%$ is added to the target variables. Figure~\ref{fig:results_noisy} and Table~\ref{tab:noisy_results} summarize the results.

Noise reduces the accuracy of all methods, but the relative advantages of ITNNR are still present. ITNNR with a single anchor remains competitive, however, ITNNR with multiple anchors and branch selection based on a learned forward model achieves substantial error reductions in all experiments.

Although the exact forward function is not available for consistency checking, it is worth comparing the performance between the exact forward function and the learned forward function. In some cases there is a significant performance difference, up to $3\times$, between the learned and exact forward function, while in other cases this difference is minimal, $1.2\times$.

Overall, ITNNR reduces inversion error by $30–85\%$ relative to kNN and by $7–57\%$ relative to MDN across noisy experiments. In these experiments the gap between MDNs and ITNNR narrows, most likely because of noisy anchors.
It might be possible to choose optimal anchors that avoid noise, as proposed in~\cite{BelaidEtAl2024WeightedPDL}. 

In summary, these results demonstrate that traditional neural networks are unsuitable for inverting non-injective functions while kNN inversion provides a reasonable baseline but requires many anchor points to achieve reasonably good accuracy. MDNs are capable of inverting non-injective functions as long as there is a discrete solution set. ITNNR reliably recovers valid inverse branches but the accuracy suffers under noisy anchors.

\section{Conclusion}
\label{sec:discussion}

This work introduces Inverse Twin Neural Network Regression (ITNNR), a deterministic machine learning framework for inverting non-injective functions. By reformulating inversion as a locally anchored regression problem, ITNNR avoids the fundamental failure modes of traditional regression methods when applied to multi-valued inverse mappings. A central insight of this work is that ITNNR predictions attached to anchors provide not only residual corrections or local linearizations. Each anchor defines a local region where a branch of the forward function is injective and hence invertible. 

The experimental results demonstrate that ITNNR reliably recovers valid inverse solutions across a wide range of settings, including low-dimensional toy problems, inverse problems with infinitely many solutions, and realistic inverse kinematics tasks. Compared to neural network, mixture density network or nearest neighbor inversion, ITNNR achieves a lower RMSE. This observation holds when the forward function is known exactly but diminishes when the problem is defined by noisy data.

For problems where input and output dimensionalities differ and infinitely many inverse solutions exist, ITNNR does not attempt to enforce uniqueness. Instead, each anchor yields one locally consistent solution. This provides a natural mechanism for biasing solutions toward preferred regions of the input space, as is commonly necessary in inverse kinematics.

To summarize, ITNNR naturally reformulates the inversion of non-injective functions from a fundamentally ill-posed global task into a collection of well-defined local inverse problems. Each of these problems can be solved by training a single neural network.

\subsection{Comments}
ITNNR can be trained effectively on data sets that can be represented by a small number of data points, since the effective training data set size is multiplied by the number of training neighbors. However, the author does not have access to large computational clusters to examine the scaling of ITNNR to more complex higher dimensional problems. 
\\\newline
For Image like problems where the data manifold is not simply connected, it might be necessary to embed the data within latent spaces of autoencoders or via contrasitve learning.
\\\newline
The code supporting this article can be found online on github~\cite{PublicInverseTwinNeuralNetworkRegression2026}.

\section{Acknowledgments}
I would like to thank Claudia Zendejas-Morales and Ziqi Xv for interesting discussions and applying this method to their student projects. 


\newpage
\appendix
\onecolumn
\section{Data sets}
\label{sec:data}

\begin{figure*}[h!]
    \centering
    \includegraphics[width=\textwidth]{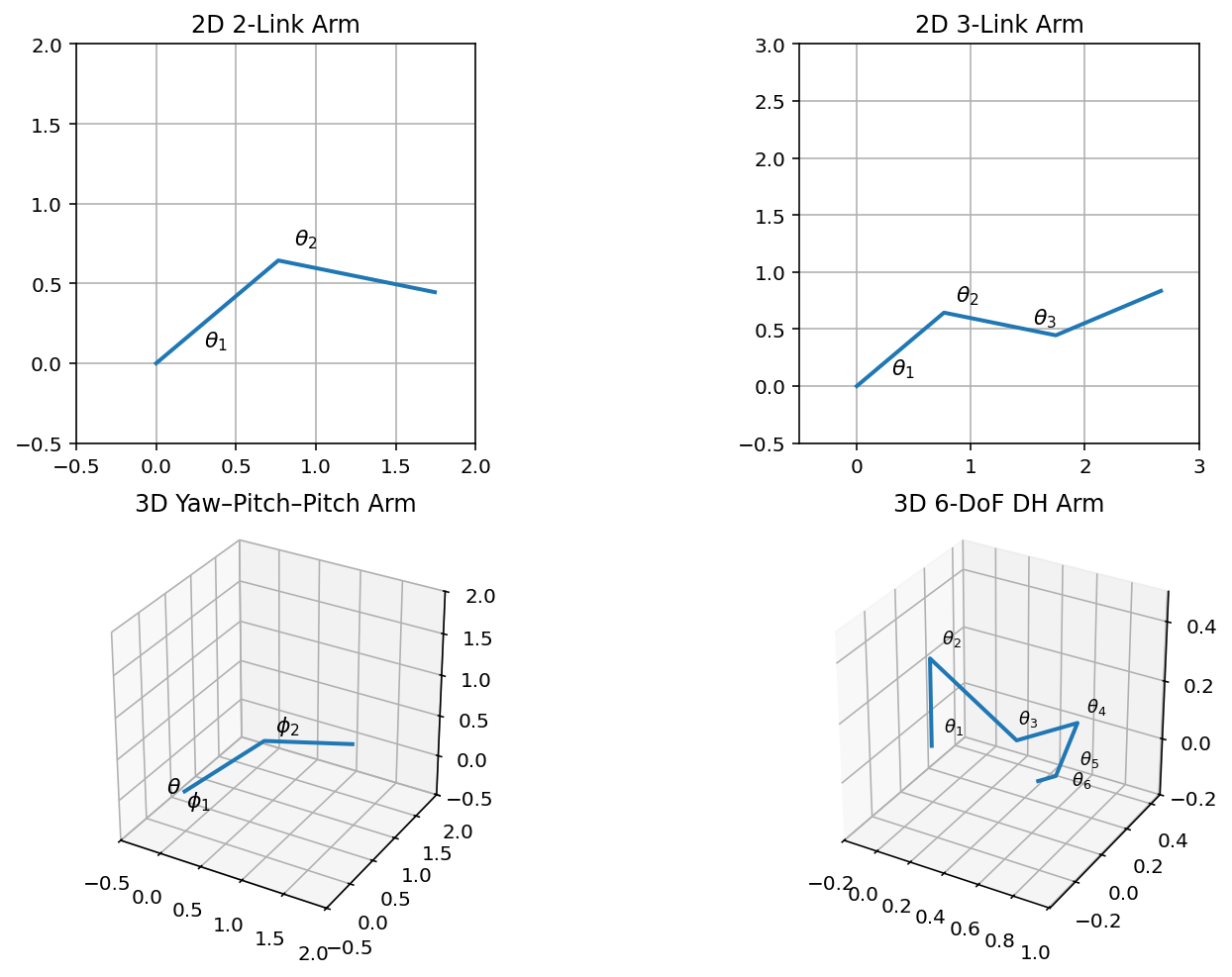}
    \caption{Robot arm geometries for Experiments 6,7,8,9.}
    \label{fig:robotarms}
\end{figure*}

The nine data sets corresponding to experiments 1-9 are generated from mathematical equations sampled from uniform distributions. Experiments 1-4 contain 500 data points, 300 anchors that are also used for training, 100 validation points and 100 test points, while experiments 5-9 contain 1000 data points, 600 anchors that are also used for training, 200 validation data points and 200 test data points. The geometry of the robot arms is visualized in figure~\ref{fig:robotarms}. 

\subsection{Exp 1: cubic polynomial}
\begin{align}
f(x)=x^3-x \ , x \in [-2,2]
\end{align}
\subsection{Exp 2: quartic polynomial}
\begin{align}
f(x)=x^4-4x^2 \ , x \in [-2,2]
\end{align}

\subsection{Exp 3: upper half of the unit ball}
\begin{align}
f(x,y)=\sqrt{1-x^2-y^2} \ , x,y \in [-1,1], x^2+y^2<1
\end{align}
\subsection{Exp 4: bivariate polynomial}
\begin{align}
f(x,y) =\left(x^{3} - 2x y^{2} + 5x + 5y,\; y^{2} - 2x y + 3x - 2y\right) \ , x,y \in [-3,3]
\end{align}
\subsection{Exp 5: trivariate polynomial}
\begin{align}
f(x,y,z) = \left( x^2 - y^2 + z,\; 2xy + z,\; x + y + z^2 \right)\ , x,y,z \in [-3,3]
\end{align}
\subsection{Exp 6: 2D 2-link planar robot arm}
\begin{align}
f(\theta_1,\theta_2) =&
\left(
L_1 \cos\theta_1 + L_2 \cos(\theta_1 + \theta_2),\;
L_1 \sin\theta_1 + L_2 \sin(\theta_1 + \theta_2)
\right)\ , \nonumber\\
&\theta_i \in [-\pi/2,\pi/2] \ , \ L_i=1
\end{align}

\subsection{Exp 7: 2D 3-link planar robot arm}
\begin{align}
f(\theta_1,\theta_2,\theta_3)=
\big(
&L_1\cos\theta_1
+ L_2\cos(\theta_1+\theta_2)
+ L_3\cos(\theta_1+\theta_2+\theta_3),\nonumber \\
&L_1\sin\theta_1
+ L_2\sin(\theta_1+\theta_2)
+ L_3\sin(\theta_1+\theta_2+\theta_3)
\big)\ , \nonumber\\
&\theta_i \in [-\pi/2,\pi/2]\ , \ L_i=1
\end{align}
\subsection{Exp 8: 3D yaw–pitch–pitch robot arm}
\begin{align}
f(\theta,\phi_1,\phi_2)=
\big(
&L_1\cos\theta\cos\phi_1 + L_2\cos\theta\cos(\phi_1+\phi_2),\nonumber\\
&L_1\sin\theta\cos\phi_1 + L_2\sin\theta\cos(\phi_1+\phi_2),\nonumber\\
&L_1\sin\phi_1 + L_2\sin(\phi_1+\phi_2)
\big)\ , \nonumber\\
&\theta,\phi_i \in [-\pi/2,\pi/2]\ , \ L_i=1
\end{align}

\subsection{Exp 9: 3D 6 degrees of freedom robot arm}
\begin{equation}
T = A_1(\theta_1) \, A_2(\theta_2) \, A_3(\theta_3) \, A_4(\theta_4) \, A_5(\theta_5) \, A_6(\theta_6) \ , \ \theta_i \in [-\pi/2,\pi/2]
\end{equation}

\noindent Denavit-Hartenberg (DH) parameters are a standard method used to model the kinematics of a robotic arm by defining coordinate frames for each joint. Each DH transformation matrix is defined as:

\begin{equation}
A_i(\theta_i) =
\begin{bmatrix}
\cos\theta_i & -\sin\theta_i \cos\alpha_i & \sin\theta_i \sin\alpha_i & a_i \cos\theta_i \\
\sin\theta_i & \cos\theta_i \cos\alpha_i & -\cos\theta_i \sin\alpha_i & a_i \sin\theta_i \\
0 & \sin\alpha_i & \cos\alpha_i & d_i \\
0 & 0 & 0 & 1
\end{bmatrix}, \quad i = 1,\dots,6
\end{equation}

\noindent with the parameters of the specific robot given by:

\[
\begin{aligned}
d &= [0.3,\, 0.0,\, 0.0,\, 0.4,\, 0.0,\, 0.1] \\
a &= [0.0,\, 0.5,\, 0.3,\, 0.0,\, 0.0,\, 0.0] \\
\alpha &= [\pi/2,\, 0,\, 0,\, \pi/2,\, -\pi/2,\, 0]
\end{aligned}
\]

\noindent The end position is extracted from the homogeneous transformation as:

\begin{equation} 
( x , y , z ) = T[:3,3]
\end{equation}

\section{Neural Network Architectures}
\label{sec:nn_architecture}
All neural networks methods, including the traditional neural network regression, the twin neural network regression methods and the mixture density networks, are built using the same architecture and implemented through the TensorFlow library \cite{Tensorflow2015}. They consist of two hidden layers with 640 neurons each and relu activation functions. The final layer contains one single neuron without an activation function, while the MDN outputs parameters of a Gaussian mixture model(weights, means and variances). I train the neural networks using the adam optimizer, and use learning rate and early stop callbacks that reduce the learning rate by 50\% or stop training if the loss stops decreasing. For this reason, it is enough to set the number of epochs large enough such that the early stopping is always triggered. The batch sizes are 32, that data is either supplied through a normal data set or through a generator that samples new training data points on the fly. For MDNs, the mixture components were adjusted in steps of 5 ranging from 5 to 500. Training of the MDNs is very unstable, which is why it was necessary to manually search for converged models.

\newpage
\sloppy

\bibliography{library}

\bibliographystyle{iopart-num}
\end{document}